\documentclass[10pt,a4paper,twoside]{article}

\usepackage[top=1in,bottom=1in,left=1.25in,right=1.25in]{geometry}
\usepackage[english]{babel}
\usepackage{bm}
\usepackage{graphicx}
\usepackage[colorinlistoftodos]{todonotes}
\usepackage{fancyhdr}
\usepackage{amsmath}
\usepackage{subfigure}
\usepackage[colorlinks,linkcolor=blue,anchorcolor=green,citecolor=red]{hyperref}
\usepackage{amsmath}
\usepackage{amssymb}
\usepackage{multirow}

\usepackage{algorithm}
\usepackage{algorithmic}

\usepackage{xspace}

\providecommand{\eg}{\textit{e.g.}\@\xspace}

\providecommand{\etal}{\textit{et al}}

\title{\rule{\textwidth}{1.5mm}\newline\newline
    \textbf{Physical Adversarial Attack on Vehicle Detector in the Carla Simulator}\\
    \begin{large}A Technical Report\end{large}\\
    \rule{\textwidth}{0.5mm}
}

\date{}
\author{
  Tong Wu \and
  Xuefei Ning \and
  Wenshuo Li \and
  Ranran Huang \and
  Huazhong Yang \and
  Yu Wang
}

\begin{document}
\maketitle
\begin{center}
  \vspace{-22pt}
  {Department of Electronic Engineering, Tsinghua University}
\end{center}

\pagestyle{fancy}
\thispagestyle{empty}
\lhead{}
\chead{}
\rhead{}
\renewcommand{\headrulewidth}{0pt}
\cfoot{Page \thepage{} of \pageref{page:EndofDoc}}

\begin{abstract}
 In this paper, we tackle the issue of physical adversarial examples for object detectors in the wild. Specifically, we proposed to generate adversarial patterns to be applied on vehicle surface so that it's not recognizable by detectors in the photo-realistic Carla simulator. Our approach contains two main techniques, an \textit{Enlarge-and-Repeat} process and a \textit{Discrete Searching} method, to craft mosaic-like adversarial vehicle textures without access to neither the model weight of the detector nor a differential rendering procedure. The experimental results demonstrate the effectiveness of our approach in the simulator.
\end{abstract}

\section{Introduction}
\label{sec:introduction}
Although deep neural networks (DNNs) have achieved impressive performance in a number of computer vision tasks, they are shown to be vulnerable to adversarial perturbations~\cite{szegedy2013intriguing} and easily misled to make wrong predictions. The existence of adversarial examples reveals a non-negligible security risk in DNN applications, including face recognition~\cite{sharif2016accessorize}, object detection~\cite{lu2017adversarial} and so on.

Adversarial attack algorithms on computer vision models can be divided into two types by their applicable domains: 1) digital attacks are conducted by directly adding imperceptible adversarial permutations to pixels of input images~\cite{szegedy2013intriguing}, while 2) physical attacks modify objects in the 3D real-world environment or physical simulators~\cite{kurakin2016adversarial,sharif2016accessorize,chen2019shapeshifter,lu2017adversarial,athalye2018synthesizing} and examine whether the permutations are physically realizable and can stay adversarial under different transformations. In this paper, we mainly focus on the latter, which is a more direct threat to visual systems in the physical world. 

Compared with the digital one, the physical attacking scenario is more challenging for the attackers since the adversarial perturbation should be robust to various environments and possibly destructive transforms. 
Previous studies that tackle the issue of physical attack vary in several key factors: attack entry point/object (\eg stickers on stop signs~\cite{lu2017adversarial,eykholt2018physical,chen2019shapeshifter}, vehicle texture~\cite{zhang2018camou}, 3D mesh norm~\cite{athalye2018synthesizing}, or universal attack~\cite{huang2020upc}), targeted task/model (\eg classification~\cite{sharif2016accessorize,athalye2018synthesizing,eykholt2017robust} and object detection~\cite{lu2017adversarial,eykholt2018physical,chen2019shapeshifter}). 
Since it is too expensive and sometimes impossible to run experiments in the real world, researchers usually conduct experiments in simulators~\cite{zhang2018camou}. However, solving the physical attacking problem in a simulator is still challenging since the rendering process is non-differentiable if one does not turn to a differentiable renderer.

Our paper attacks a vehicle detector by manipulating the vehicle texture in a photo-realistic simulator, and the mosaic-like adversarial patterns crafted by our method are constructed by regular and colorful blocks. Specifically, we propose a discrete search method to craft the adversarial texture pattern. 
Also, we carefully construct the adversarial pattern via an \textit{Enlarge-and-Repeat} process, and this process narrows down the pattern search space, which makes it efficient to apply discrete search. 
Using the proposed \textit{Enlarge-and-Repeat} process and the discrete search attacking method, we achieve remarkable attacking results in physical simulators, and we highlight that easy mosaic-like adversarial patterns can be a non-negligible threat to modern vehicle detectors. 

\section{Related Studies}
\label{sec:related}
\noindent\textbf{Physical Attack}
Adversarial attacks in the real world are first observed by~\cite{kurakin2016adversarial} where they found that images with adversarial permutation printed out on paper remain effective.
There have been many physical attacking studies that attack classifiers: Sharif~\cite{sharif2016accessorize} generated glasses with patterns to fool facial recognition system; 
Athalye \etal~\cite{athalye2018synthesizing} proposed Expectation-over-transformation (EoT) and constructed physical attacks by with 3D-printed objects; 
Evtimov~\cite{eykholt2017robust} used black and white stickers to attack stop signs.
Lu~\etal~\cite{lu2017standard} pointed out that these methods fail to attack object detectors~\cite{redmon2017yolo9000,ren2015faster}, which is a more challenging task. 
Later, several approaches~\cite{lu2017adversarial,eykholt2018physical,chen2019shapeshifter} claimed that detectors are also fragile if some modifications to the loss functions are made, and region proposals are taken into consideration. The attacks are also performed by adding permutations to stop signs, which has a planar surface.
While most recently, Zhang~\etal~\cite{zhang2018camou} used a clone network to attack Mask R-CNN~\cite{he2017mask} with camouflages painted to the vehicle surface in the CARLA simulator~\cite{dosovitskiy2017carla}; And Huang~\etal~\cite{huang2020upc} proposed a universal physical camouflage attack which can be camouflaged as texture patterns on object surfaces such as human accessories/car paintings. 


Our work is most closely related to~\cite{zhang2018camou} in that we all craft adversarial texture of vehicles, 
but we perform a more careful construction of the adversarial pattern search space, and employ a discrete search method instead of using a ``clone model'' to mimic the black-box behavior. We introduce our experiences of reproducing their clone model solution and discuss its limitation in Sec.~\ref{subsec:previous}.

\noindent\textbf{Object Detection}
Object detection based on deep learning has long been an important task and has aroused comprehensive interest. Modern methods include anchor-based and anchor-free~\cite{law2018cornernet, duan2019centernet} detectors, and the former can be further divided into two-stage~\cite{lin2017feature,ren2015faster,li2017lighthead,he2017mask} and one-stage~\cite{redmon2017yolo9000, liu2016ssd} methods based on whether region proposals are generated. In this paper, we use an light-headed RCNN~\cite{li2017lighthead} as the targeted model. 
\section{Methodology}
\label{sec:methodolody}
Our goal is to attack an object detector by rendering camouflages all over the car and mislead the detector so that it would fail to produce bounding boxes with high IoU or misclassify the car.
In this section, we present our attacking algorithm in two parts: 1) we first explain some details of our physical simulation including rendering strategy and generation of training data; and then 2) we propose a discrete searching algorithm to perform the black-box attack without access to neither the model weight of the detector nor a differential rendering procedure.

\subsection{Physical Simulation and Camouflage Rendering}
\label{subsec:render}

\begin{figure}[t]
\centering
	\begin{minipage}{0.68\textwidth}
	\centering  
	\includegraphics[width=0.95\linewidth]{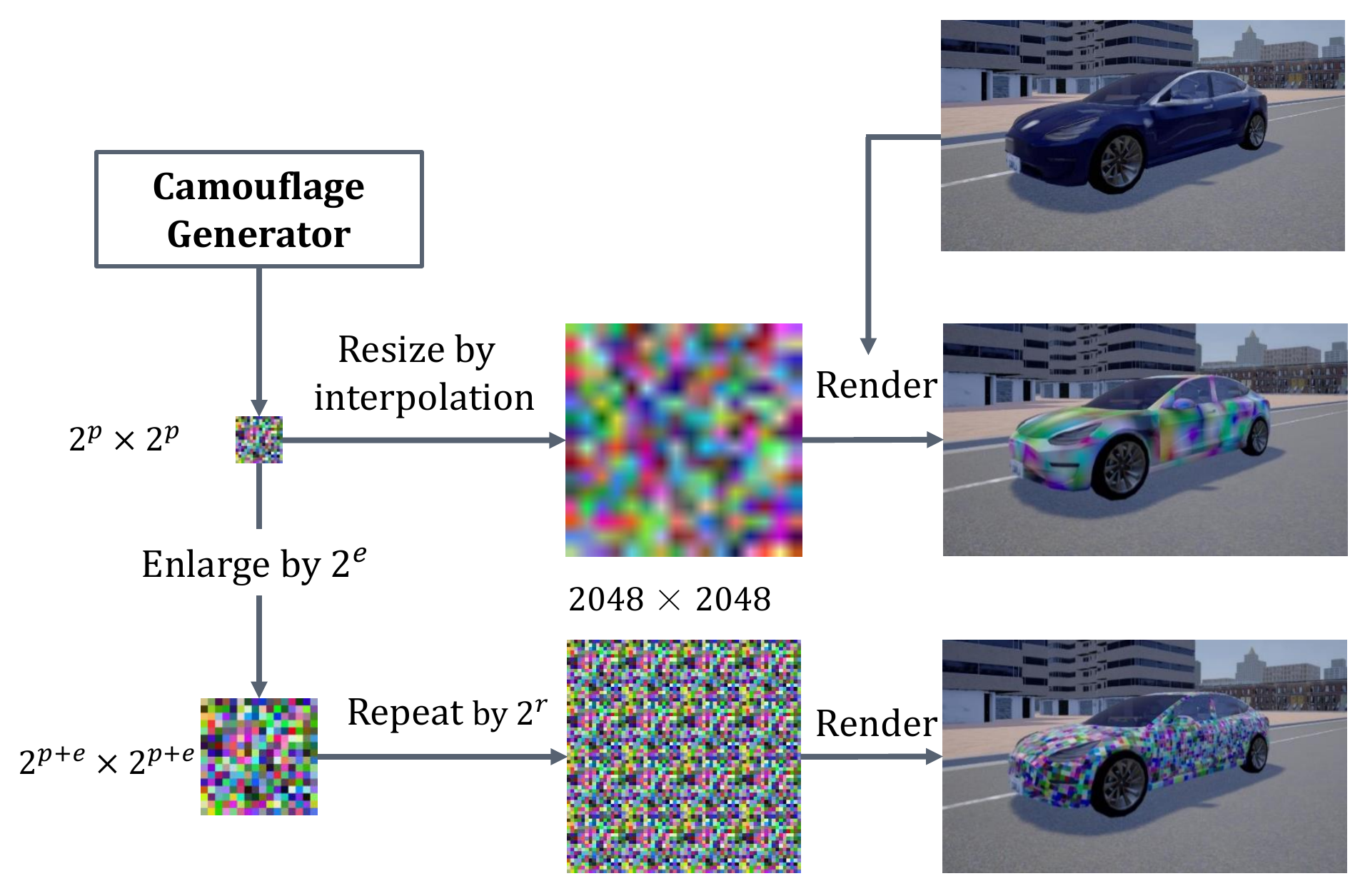} 
    \caption{
        Two ways of generating a camouflage from a pattern. 
        Upper: Directly resize the pattern by interpolation. 
        Lower: Our Enlarge-and-Repeat manner.
    }
    \label{fig:resolution}
	\end{minipage}
	\begin{minipage}{0.28\textwidth}
	\centering
	\includegraphics[width=0.95\linewidth]{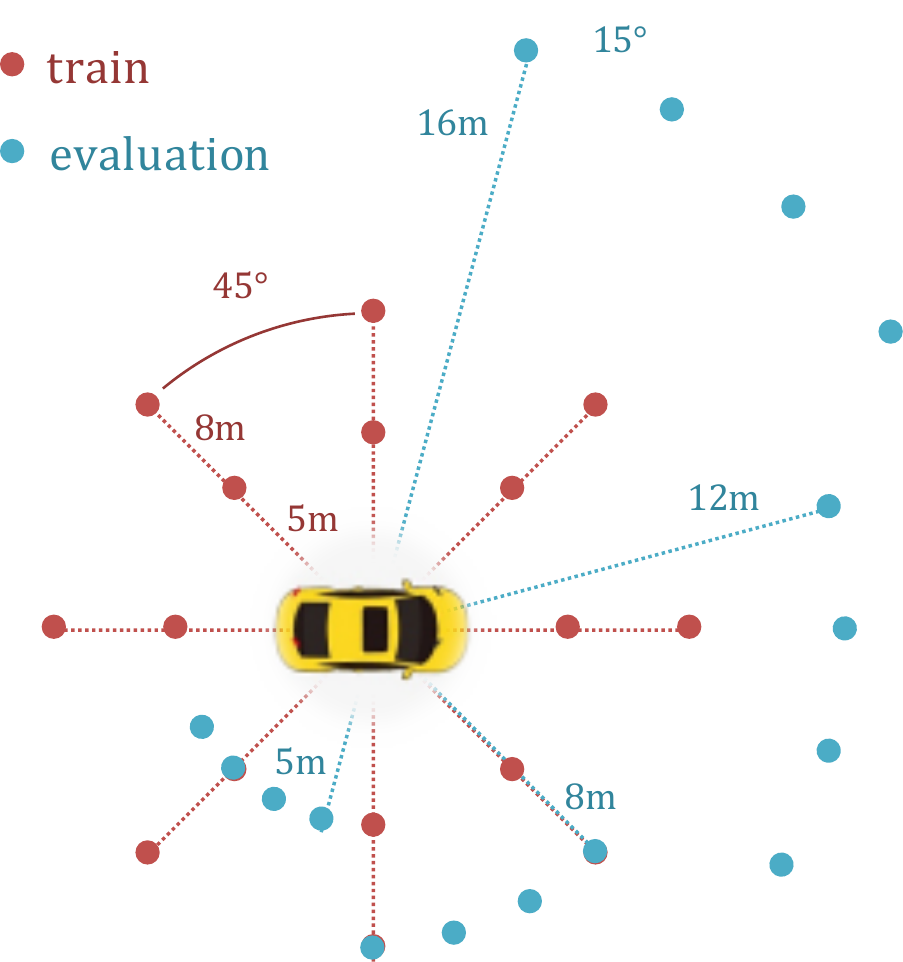} 
    \caption{
        The red and blue points indicate the transformations at training and testing time, respectively
    }
    \label{fig:direction}
	\end{minipage}

\end{figure}


We leverage the open-source simulator CARLA~\cite{dosovitskiy2017carla} to conduct the experiments. 
The CARLA simulator is built on Unreal Engine 4, and provides various maps, vehicle models, and enable one to use simple APIs to manipulate the environment, vehicle state, vehicle texture, camera state and so on. 
Given a piece of camouflage pattern, which is of the size of $2048\times2048$\footnote{The size of the vehicle textures is $2048\times 2048$ in the CARLA simulator.}, it can be rendered onto the vehicle surface with Unreal API. The camera sensor we create would receive rendered 2D RGB pictures of the scene, including the camouflaged vehicle. By changing the camera transformations (various distance and viewpoints, in our setting), we generate image data for both training and evaluation.

If the size of the pattern image is smaller than the default configuration above, it would be resized via interpolation first, which may induce blurring to the original pattern, especially when the initial size is rather small. Our experimental results show that the attacking performance degrades when the pattern is highly blurred.

In this paper, we explore the design space of the adversarial camouflage pattern, by adopting different \textit{Enlarge-and-Repeat (ER)} configurations. 
Specifically, assume the pattern is a $2^p \times 2^p$ pixels RGB square block, which is much smaller than the whole camouflage texture.
To construct the camouflage texture, instead of directly resizing the pattern unit with interpolation, we first enlarge each pixel to a $2^{e} \times 2^{e}$ pixels block and get a bigger pattern with the length of $2^{p+e} \times 2^{p+e}$; and then, it's repeated $2^r$ times both vertically and horizontally to get the final camouflage texture, a square of length $2^{p+e+r}=2048$. 
The ER process takes two factors into consideration.
First, in order to make the patterns more conspicuous and robust towards various environmental factors, one can \textbf{enlarge} each pixel into a small square block. 
Second, since the camouflage is rendered onto the whole body except for the windows and tyres, so from different viewing angles, only part of the pattern can be caught by camera. Thus, a natural idea is to \textbf{repeat} the same adversarial pattern to make the best use of its adversarial characteristics.
Overall, it seems to be painting a mosaic-like camouflage on the vehicle.

The transformation used in our paper would be elaborated in Sec.~\ref{sec:experiments}.


\begin{figure}[tb]
\centering
    \includegraphics[width=0.95\linewidth]{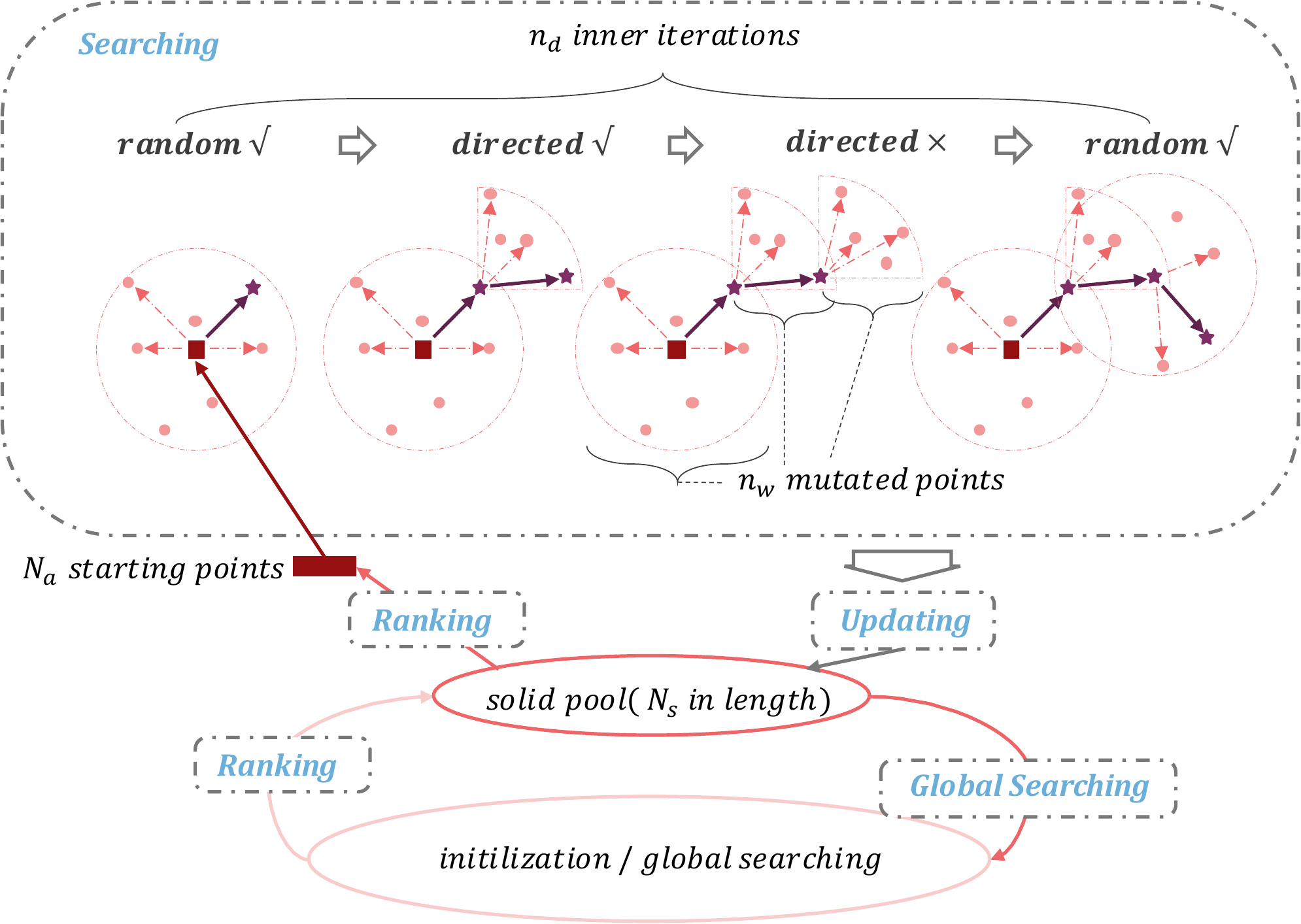} 
    \caption{Discrete searching illustration. The searching process from left to right gives examples of the two mutation strategies, namely randomly and directed. It depends on whether it is the initialization process and whether the best mutated point outperforms the current point}
    \label{fig:searching}
\end{figure}

\subsection{Discrete Searching}
\label{subsec:search}
Since the rendering process is generally non-differential, it's not applicable to update the adversarial pattern using back-propagated gradients. Also, we assume no knowledge of the vehicle detection model, and aim at black-box attacking the vehicle detection model using only queries.
We conduct the black-box attack by iteratively refining the camouflage using a mutation-based search method. 
Note that in~\ref{subsec:render} we have constrained the searching space to a small patch of pattern instead of the full car surface to increase feasibility. We'll only use camouflage, i.e., patterns after the ER process, to explain the searching algorithm here. The notations and concrete description of the algorithm are summarized in Alg.~\ref{alg:searching}, and a graphical illustration is shown in Fig.~\ref{fig:searching}.

Specifically, we first initialize $N_c$ randomly generated camouflages and select the top $N_a$ of them to be the starting points of the searching. In each inner iteration, $n_w$ mutated points are generated and evaluated. And there are two mutation strategies: 1) scattering inside an epsilon-ball of $l_\infty$ with random direction and step sizes; 2) directed update along a preferred direction with randomized step sizes. In the first iteration, there is no preferred direction at every starting point, thus we adopt the randomized scattering mutation strategy, and find the best mutated point among the $n_w$ mutated point. Then the preferred direction $\delta$ is determined by the current point $C_i, i=0,...,n_w-1$ and the best mutated point $\hat{C}_i$ (Line 10, Alg.~\ref{alg:searching}), and the best mutated point is chosen as the new current point (Line 11, Alg.~\ref{alg:searching}). Then in the next inner iteration, directed update mutation along the preferred direction would be adopted. 
In this way, these two mutation strategies are interleaved. The randomized scattering mutation is more localized and is finding a direction, and the directed update enables faster exploration of the search space. One inner loop contains $n_d$ iterations of the aforementioned updates, and the total search process contains $n_r$ inner loops.


Meanwhile, we also maintain a set of solid camouflages with the best attacking performances, namely \textit{solid pool}, which would be used to generate the starting points of the next inner loop. And a larger epsilon $\epsilon_2$ and no direction guidance is used in the starting points generation of each inner loop (Line 21, Alg.~\ref{alg:searching}). This global search intends to enhance the exploration to a larger $l_\infty$ norm ball to avoid being stuck in a local minimum.


During the search process in Alg.~\ref{alg:searching}, we evaluate each camouflage pattern by 1) Assigning the camouflage pattern as the vehicle texture; 2) Capturing the rendered 2D RGB images using camera sensors with different camera transformations; 3) Running the vehicle detector on the captured images and calculating the average detection score.

\begin{algorithm}[h]
\caption{Discrete searching to find adversarial patterns}
\label{alg:searching}
\begin{algorithmic}[1]
    \REQUIRE ~~\\ 
    Randomly initialize $N_c$ camouflages in the search space $\mathbb{R}^d, d=H\times W\times3\times256$.
    \ENSURE ~~\\ 
    A set of solid camouflages that gains the best attacking performance.

    \STATE Select the top $N_s$ camouflages with the lowest average score to initialize the \textit{solid pool}
\FOR{each $j_r \in\{0,...,n_r-1\}$}   
    \STATE Choose top $N_a (N_a<N_s)$ camouflages from the \textit{solid pool} to be the starting points, namely $C_i, i \in\{0,...,N_a-1\}$.
    \FOR{each $i \in\{0,...,N_a-1\}$}     
        \STATE Initialize the diretion $\delta$ to be a random matrix of ${1, -1}^{H\times W}$.
        \FOR{each $j_d \in\{0,...,n_d-1\}$}   
            \STATE Take $n_w$ random searching steps with $C_i^{j_w} = Clip(C_i + Random(H,W,3)~\cdot~\epsilon_1~\cdot\delta), j_w=0,...,n_w-1$
            \label{step:search_local}
            \STATE Select $\hat{C_i}$, which is the best among the $C_i^{j_w}, w=0,...,n_w-1$
            \IF{$\hat{C_i}$ is better than $C_i$}
                \STATE Estimate a rough local direction $\delta=sign(\hat{C_i}-C_i)$
                \STATE Update $C_i = \hat{C_i}$
            \ELSE
                \STATE Take direction $\delta$ to be a random matrix in $\{1, -1\}^{H\times W}$, 
                \STATE $C_i$ remains unchanged
            \ENDIF    
        \ENDFOR 
    \ENDFOR
    
    \STATE Update \textit{solid pool} by ranking the original set with all the new ones generated during searching
    \FOR{each $i \in\{0,...,N_s-1\}$}     
        \STATE Take $n_g$ random searching steps with $C_i^{j_g} = Clip(C_i + Random(H,W,3)~\times \epsilon_2), j_g=0,...,n_g-1$
        \label{step:search_global}
    \ENDFOR
    \STATE Again update \textit{solid pool} by ranking the original set with all the new ones generated during searching
\ENDFOR
\end{algorithmic}
\end{algorithm}

\section{Experiments}
\label{sec:experiments}
\subsection{Experiment Setup}
\noindent\textbf{implementation Details.}
The hyper-parameters in Algorithm~\ref{alg:searching} are as follows: 
we set $N_c=100, N_s=20$ and $N_a=5$ for the size of initialization, solid pool, and searching starting points, respectively. 
$n_w$ is set to 20. The inner and outer iteration numbers, $n_d$ and $n_o$, are set to $3$ and $5$, respectively. The decreasing of average detection score usually gets quite slow after 5 outer iterations, and thus we get it stopped.
As for the mutation parameters, we set $\epsilon_1=5$ and $\epsilon_2=10$.
We use Light-Head RCNN~\cite{li2017lighthead} pretrained on MSCOCO~\cite{coco} as the targeted object detector.
During training, we use 16 camera transformations: the distances are set as 5 and 8 meters away from the vehicle, and 8 uniformly distributed viewing angles are used for each of the distances.
During testing, we use 96 camera transformations with the distances set as 5, 8, 12 and 15 meters and 24 uniformly distributed viewing angles for each of them. Different transformations are shown in Fig.~\ref{fig:direction}.


\noindent\textbf{Evaluation Metrics}
One of the metrics we use is~\textit{average prediction score} $S_{avg}$, where we take the highest box prediction that belongs to the car category, and average the results under various pre-defined camera transformations.
Since the detector considers no car detected when the highest prediction score is lower than a threshold (set as 0.5), we follow UPC~\cite{huang2020upc} to use~\textit{precision with a threshold of 0.5} $P_{0.5}$ as another metric. Our definition is slightly different, by directly calculate the the ratio between successfully detected vehicles images and the total number.

\noindent\textbf{Baselines}
Following~\cite{zhang2018camou}, we compare with two baseline cases, namely \textit{clean} and \textit{random}.
We choose \textit{black} as a standard vehicle color for the \textit{clean} case, and we randomly generate 100 camouflages in the same manner as the initialization part in Algorithm~\ref{alg:searching} for the \textit{random} case.
\footnote{Due to the lack of implementation details in their paper~\cite{zhang2018camou}, and our implementation of their method didn't work well, so we didn't include it as a baseline.}

\subsection{Experimental Results}

We first evaluated the influence of different settings of the \textit{\textbf{E}nlarge-and-\textbf{R}epeat} process, where we would use abbreviations for convenience, e.g., E$e$-R$r$ means first enlarging each pixel of the pattern for $2^{e}$ times and then repeating it for $2^{r}$ times, which is presented more concretely in Sec.~\ref{subsec:render}. 
As shown in Tab.~\ref{table:res}, the detector can detect a clean car with $S_{avg}=0.89$ and $P_{0.5}=0.91$.
With randomly generated camouflages, we can see that the attacking effect varies under different ER settings. Specifically, E5-R2 with a medium pattern frequency performs the best with a drop of 0.33 and 0.34 from the clean baseline for $S_{avg}$ and $P_{0.5}$, respectively. The attacking effect gets less remarkable as the frequency gets either higher (E4-R3) or lower (E7-R0). The visualization of their comparison is shown in Fig.~\ref{fig:resolution}.
We observed that the two metrics share the same tendency. 

And then we test the effectiveness of our searching strategy, as shown in Tab.~\ref{table:res}. With the same ER setting, compared with the random baseline, $S_{avg}$ and $P_{0.5}$ further decrease for $[0.09, 0.13]$ and $[0.08, 0.20]$ points, respectively. 
Overall, the best attacking results are achieved by applying E5-R2 and our discrete searching method. Specifically, a drop of 0.46 and 0.52 are observed in $S_{avg}$ and $P_{0.5}$, respectively.

Notice that the decreasing is quite remarkable even before the searching stage with randomly initialized camouflages, indicating the contribution of the \textit{Enlarge-and-Repeat} process itself, which will be further evaluated in Sec.~\ref{subsec:ablation}.

\begin{table}[ht]
    \center
    \begin{tabular}{|c|c|c|c|c|c|c|}
    \hline
    \multirow{2}{*}{Method} & \multicolumn{3}{c|}{$S_{avg}$} & \multicolumn{3}{c|}{$P_{0.5}$}\\ \cline{2-7} 
    & \multicolumn{1}{l|}{clean} & \multicolumn{1}{l|}{random} & our  & \multicolumn{1}{l|}{clean} & \multicolumn{1}{l|}{random} & our \\ \hline
    E7-R0 & \multirow{4}{*}{0.89}   & 0.73 (-0.16) & 0.64 (-0.25) & \multirow{4}{*}{0.91}   & 0.75 (-0.16) & 0.67 (-0.24) \\ \cline{1-1} \cline{3-4} \cline{6-7} 
    E6-R1 & & 0.68 (-0.21) & 0.56 (-0.33) &  & 0.71 (-0.20) & 0.51 (-0.40)\\ \cline{1-1} \cline{3-4} \cline{6-7} 
    E5-R2 & & 0.56 (-0.33) & \textbf{0.43 (-0.46)} & & 0.57 (-0.34) & \textbf{0.39 (-0.52)} \\ \cline{1-1} \cline{3-4} \cline{6-7} 
    E4-R3 & & 0.60 (-0.29) & 0.49 (-0.40) &  & 0.59 (-0.32) & 0.47 (-0.44) \\ \hline
    \end{tabular}
    \caption{Experimental results of average prediction scores and precision}
    \label{table:res}
\end{table}
 
\begin{figure}[t]
    \centering
    \includegraphics[width=\linewidth]{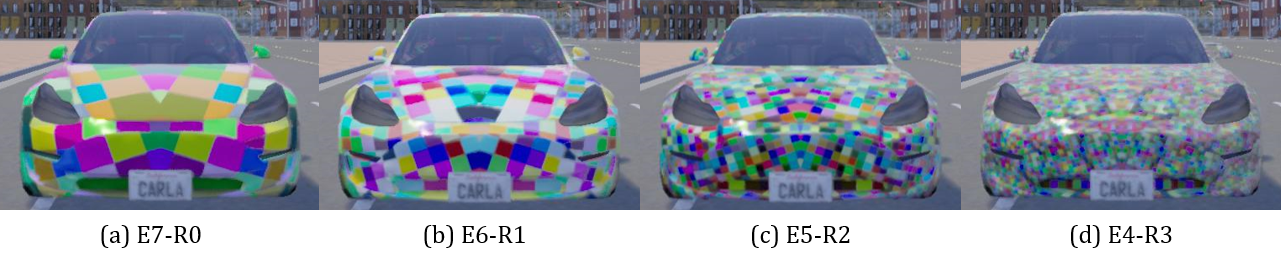} 
    \caption{Visualization of rendering results under four different ER settings}
    \label{fig:resolution}
\end{figure}

\subsection{Ablation Study}
\label{subsec:ablation}
\noindent\textbf{Comparison of different rendering methods}
We further compared the attacking effect of different rendering methods in Tab.~\ref{table:render}. Specifically, we compared the ER process with the basic resizing via interpolation. 
We observed a huge gap between these two methods, and the former has a rather weak attacking performance compared with the clean case. This might be due to that the camouflages produced by directly resizing have a continuous color pattern and highly blurred boundaries. And this pattern is closer to a normal car painting, thus more likely to be correctly recognized.

\begin{table}[]
    \center
    \begin{tabular}{cccllll}
    \cline{1-3}
    \multicolumn{1}{|c|}{Methods} & \multicolumn{1}{c|}{Avg Score} & \multicolumn{1}{c|}{$P_{0.5}$} &  &  &  &  \\ \cline{1-3}
    \multicolumn{1}{|c|}{Clean} & \multicolumn{1}{c|}{0.89} & \multicolumn{1}{c|}{0.91} &  &  &  &  \\ \cline{1-3}
    \multicolumn{1}{|c|}{Bi-linear-Random} & \multicolumn{1}{c|}{0.85} & \multicolumn{1}{c|}{0.88} &  &  &  &  \\ \cline{1-3}
    \multicolumn{1}{|c|}{ER-Random} & \multicolumn{1}{c|}{0.56} & \multicolumn{1}{c|}{0.57} &  &  &  &  \\ \cline{1-3}
    \multicolumn{1}{|c|}{ER-Search} & \multicolumn{1}{c|}{\textbf{0.43}} & \multicolumn{1}{c|}{\textbf{0.39}} &  &  &  &  \\ \cline{1-3}
    \multicolumn{1}{l}{} & \multicolumn{1}{l}{} & \multicolumn{1}{l}{} &  &  &  & 
    \end{tabular}
    \caption{Experimental results of different rendering methods}
    \label{table:render}
\end{table} 

\noindent\textbf{Visualization and case study}
Here we analyze some typical cases of detector behaviors under attack, and we roughly classify them by the quality of bounding box produced by Region Proposal Network (RPN) that finally owns the highest score for the car category:
1) In some cases, the bounding box covers the target well with a lower prediction score; 
2) Sometimes the bounding box only partially covers the target with a further lower prediction score;
3) The best attacking case is that the detector totally misses the target, and even the background elements have a higher prediction score. Actually, the detector considers no car detected here when the highest prediction score is lower than a threshold, usually set as 0.5.

\begin{figure}[t]
    \centering
    \includegraphics[width=\linewidth]{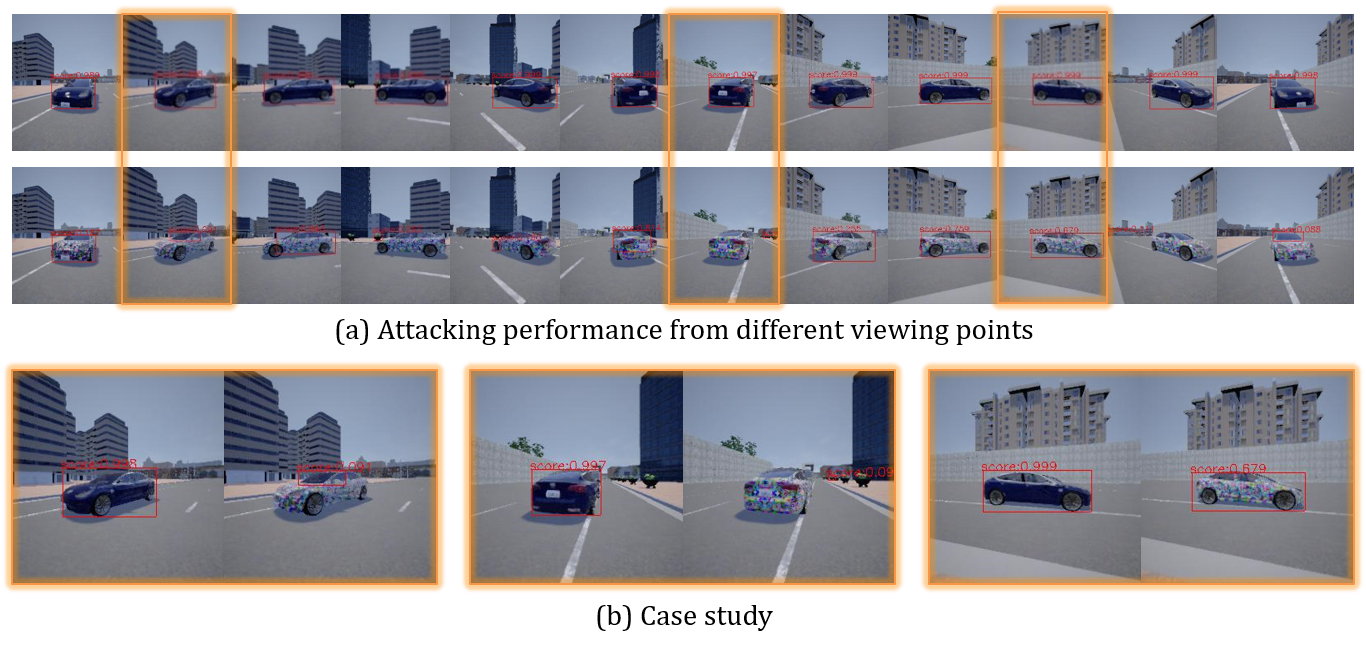} 
    \caption{
        Visualization of detector behaviors under different camera transformations and three typical cases of bounding box
    }
    \label{fig:case}
\end{figure}

\subsection{Discussion on Re-implementation of Previous Approach}
\label{subsec:previous}
Since our work is inspired and closely related to Zhang~\etal~\cite{zhang2018camou}, we will discuss about their method briefly.
They use a clone network to mimic the joint behavior of the simulator and the detection model. And they alternatively train the clone network and craft the adversarial texture using the gradients of the clone network.
Since neither the code nor the implementation detailsf is provided, we try to implement their solution ourselves. However, we found that clone network does not work well in our re-implementation. The clone network only mimics the simulator and detection model well in a small scope of color space, while failing to provide meaningful gradients after few steps of adversarial updates. Thus, the attacking steps are only attacking the clone network, instead of the joint system of the simulator and the detection model. 

\subsection{Future Work}
Currently, the object detector is pre-trained on standard real-world datasets~\cite{imagenet,coco} with only clean examples. It is worthy of exploring whether utilizing a noisy synthesized vehicle dataset with random surface patterns as extra training sources would increase the model's robustness. It shares a similar idea with the widely used adversarial training~\cite{madry2017towards} in digital approaches.
\section{Conclusion}
\label{sec:conclusion}
In this paper, we propose an \textit{Enlarge-and-Repeat} process and a discrete search method to craft physically robust adversarial textures for attacking vehicle detection models. Our experiments are carried out in a photo-realistic simulator. And the experimental results show that the proposed attacking method successfully fools the vehicle detector from multiple orientations and distances, and both techniques are effective in crafting stronger adversarial texture.



\section*{Acknowledgments}
The authors gratefully acknowledge the support from TOYOTA.









\clearpage
{
\small
\bibliographystyle{ieeetr}
\bibliography{references}
}

\label{page:EndofDoc}
\end{document}